\documentclass{egpubl}
\usepackage{egsr2021}
\usepackage{amsmath,bm}
\usepackage{booktabs}
\JournalSubmission    
\usepackage[T1]{fontenc}
\usepackage{dfadobe}  

\usepackage{cite}  
\BibtexOrBiblatex
\electronicVersion
\PrintedOrElectronic
\ifpdf \usepackage[pdftex]{graphicx} \pdfcompresslevel=9
\else \usepackage[dvips]{graphicx} \fi

\usepackage{egweblnk} 

\usepackage{ulem}
\usepackage{amsfonts}

\title{Deep Portrait Lighting Enhancement with 3D Guidance}

\author[Fangzhou Han, Can Wang, Hao Du \& Jing Liao]
{\parbox{\textwidth}{\centering Fangzhou Han$^\dag$\orcid{0000-0002-3641-1667}, Can Wang\thanks{equal contribution}\orcid{0000-0002-5102-1464}, Hao Du\orcid{0000-0003-1330-4532} and Jing Liao\thanks{corresponding author}\orcid{0000-0001-7014-5377} 
        }
        \\
{\parbox{\textwidth}{\centering City University of Hong Kong
      }
}
}

\begin{document}


\teaser{
  \centering
  \includegraphics[width=\linewidth]{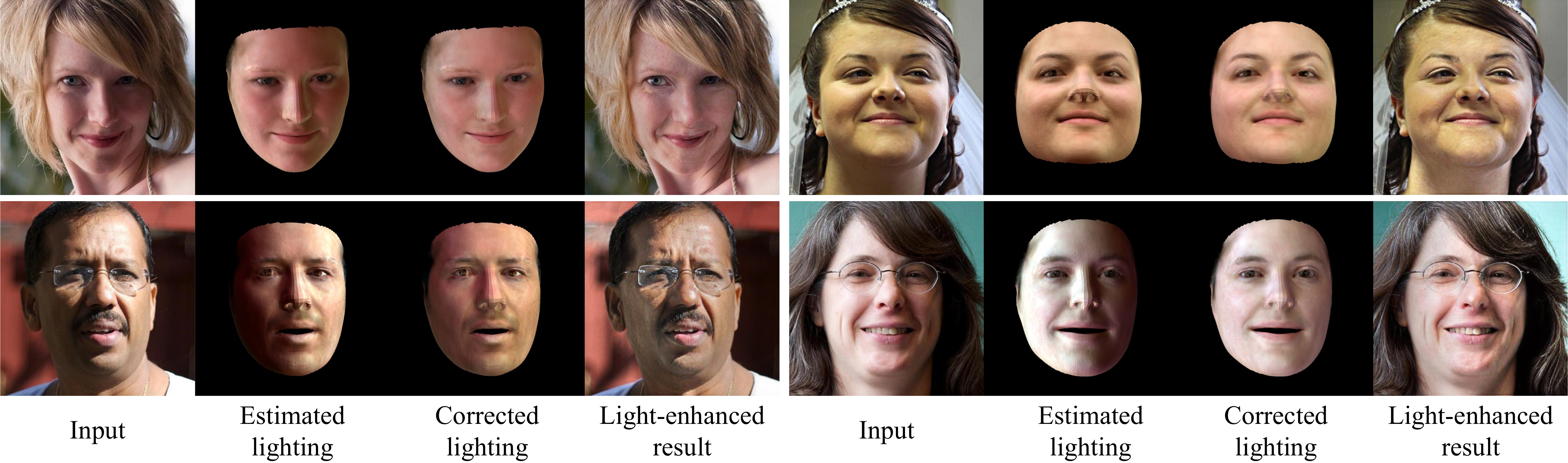}
  \caption{Portrait lighting enhancement results of our method. With 3D guidance, our method enables producing photo-realistic lighting-enhanced images.  }
  \label{fig:teaser}
}

\maketitle

\begin{abstract}
Despite recent breakthroughs in deep learning methods for image lighting enhancement, they are inferior when applied to portraits because 3D facial information is ignored in their models. To address this, we present a novel deep learning framework for portrait lighting enhancement based on 3D facial guidance. Our framework consists of two stages. In the first stage, corrected lighting parameters are predicted by a network from the input bad lighting image, with the assistance of a 3D morphable model and a differentiable renderer. Given the predicted lighting parameter, the differentiable renderer renders a face image with corrected shading and texture, which serves as the 3D guidance for learning image lighting enhancement in the second stage. To better exploit the long-range correlations between the input and the guidance, in the second stage, we design an image-to-image translation network with a novel transformer architecture, which automatically produces a lighting-enhanced result. Experimental results on the FFHQ dataset and in-the-wild images show that the proposed method outperforms state-of-the-art methods in terms of both quantitative metrics and visual quality. We will publish our dataset along with more results on \url{https://cassiepython.github.io/egsr/index.html}.

\ccsdesc[500]{Computing methodologies~Computational photography}
\ccsdesc[500]{Computing methodologies~Image processing}

\printccsdesc   
\end{abstract}  
\section{Introduction}
\label{sec:intro}

In portrait photography, lighting is one of the most important elements for photo quality and aesthetics. Professional portrait photographers can capture desirable illumination of their subjects by using specialized equipment, such as flashes and reflectors. However, most casual photographers are not able to create such compelling photographs and their photos taken in poor lighting conditions may suffer from overexposure, underexposure, low contrast, and high ISO noise problems. To better emulate professional-looking portraits, we aim to develop a deep learning method for enhancing the lighting of portrait photos taken by casual users. 

Previous deep learning methods for lighting enhancement fall into two categories. One is to learn an image-to-image translation network from the source distribution of bad lighting images to the target distribution of enhanced lighting images. Generative adversarial networks (GANs) are often served as the backbone networks because of their excellent performance in synthesizing images ~\cite{jiang2019enlightengan,wang2019underexposed}. Another category is to reformulate this task as an image-specific curve estimation problem~\cite{yuan2012automatic,guo2020zero}. By learning the parameters of curve functions from input images, pixel-wise adjustments can be made to obtain enhanced results. Despite their major progress on improving the overall brightness of input images, these generic methods fail to generate delicate lighting enhancement effects on portraits. This is because these methods are limited in the 2D image domain, ignoring vital 3D information, such as face geometry and lighting directions.  


To fully take advantage of the 3D information, we present a 3D-guided portrait lighting enhancement framework with two stages. In the first stage, a \textbf{lighting parameter correction network} estimates the original lighting parameters from the input image and predicts the corrected parameters to adjust the original lighting as well. A differentiable renderer~\cite{liu2019soft} is adopted to render the reconstructed 3D morphable model (3DMM) of the input with lighting parameters, while enabling the network to receive supervision from 2D images. As a result, a rendered portrait image with corrected lighting is obtained to serve as 3D guidance in the second stage.
In the second stage, an \textbf{image enhancement network} learns a mapping from the bad lighting image distribution to the enhanced lighting image distribution, conditioned on the 3D guidance provided earlier. The guidance image, containing shading and texture information under the corrected lighting, is greatly helpful for image enhancement, and the major challenge is how to exploit semantically meaningful correlations between the input image and the guidance. Convolution Networks~(CNN) models are less effective in this task as the local inductive priors of convolution operation make it difficult to exploit long-range correlations. In contrast, the transformer architecture ~\cite{vaswani2017attention} abandons the baked-in local inductive prior and thus supports global interactions via the dense attention module. Therefore, we adopt the transformer structure in our image enhancement network. Unlike many visual transformer tasks that only consider self-correlations in the input~\cite{carion2020end,dosovitskiy2020image}, we model the correlations between the input and the guidance by setting the guidance as query and the input as key. This helps recover missing details of the input, especially in underexposure and overexposure regions, as missing details can be borrowed from other parts of the face by leveraging long-range correlations.

Besides leveraging 3D information, another challenge of portrait lighting enhancement is how to collect bad-and-enhanced lighting image pairs to support training. To the best of our knowledge, current public image light enhancement datasets~\cite{wei2018deep,cai2018learning} are for scenes rather than for portraits. Though some face relighting datasets have been proposed~\cite{gross2010multi, zhou2019deep}, their data contains portraits with various lighting positions yet lacks explicit definition of enhanced or desirable lighting, which is inappropriate for the task. Furthermore, collecting real data is expensive and laborious as it requires a large number of identities as well as professional light stage system \cite{murmann2019dataset}.
To this end, we propose an efficient data synthesis method. We first ask volunteers to vote for visually pleasant images taken under good lighting conditions from the FFHQ dataset~\cite{karras2019style} as our target images. To create training pairs, each target image is then degraded to a bad lighting condition by extrapolating its lighting parameters away from the center of the target dataset, 
followed by post-processing and manual screening. We demonstrate the model trained on our synthesized dataset generalizes well on in-the-wild images.

Our technique contributions are summarized as follows:
\begin{itemize}

\item We design a two-stage framework for portrait lighting enhancement with 3D guidance, which outperforms previous 2D approaches by exploiting 3D facial information.
\item A transformer-based image-to-image translation network is proposed to model long-range correlations between the guidance image and the input image, which further improves the performance of our method in lighting enhancement.
\item We propose an efficient data synthesis method that can produce photo-realistic training pairs for the portrait lighting enhancement task. This dataset will be released for research purposes. 
\end{itemize}

\section{Related Work}

\subsection{Image Enhancement}

Image enhancement aims to beautify images based on certain standards, such as from low-light to normal-light or from low dynamic range to high dynamic range.
A series of methods use reinforcement learning \cite{hu2018exposure,yu2018deepexposure} or GAN \cite{deng2018aesthetic} to learn the best adjustment parameters, including brightness, contrast, curves, etc. For example, Hu~\emph{et al.}\cite{hu2018exposure} consider these basic image exposure adjustment operations as basic action units, and use reinforcement learning to solve for the optimal action sequence and parameters of each action to adjust the exposure of the input image. On the basis of \cite{hu2018exposure}, Yu~\emph{et al.}\cite{yu2018deepexposure} use semantical masks to adjust the exposure by regions.
EnhanceGAN~\cite{deng2018aesthetic} uses GAN for weakly supervised image enhancement. Its generator performs a global color adjustment, and the discriminator determines whether the results are enhanced images or not.
This kind of methods is explainable and fast, yet they can only adjust the brightness of pixels, and cannot recover missing details in the underexposure or overexposure regions.

The second series of methods, such as \cite{chen2018deep, zhou2020gan} directly learn the mapping between the two image domains and generate the enhanced target image by image-to-image translation networks. Chen~\emph{et al.}\cite{chen2018deep} use a two-stage GAN similar to CycleGan \cite{zhu2017unpaired} to perform image enhancement. The generator uses the structure of U-Net~\cite{ronneberger2015u} and takes global features as input to reveal high-level information and to determine local adjustments for individual pixels. Comparing to learning curves or parameters, directly generating images can enhance images with greater flexibility.

These two categories of methods are limited in the 2D image domain without considering 3D information. So they are feasible when enhancing the overall brightness or color tone of generic images, but it would lead to unnatural effects when enhancing the lighting of faces, which are sensitive to 3D shapes. In our method, 3D facial guidance is involved and proved useful.
\subsection{Face Relighting}
Portrait relighting aims to change the lighting condition of face images given a group of arbitrary lighting parameters.
A series of methods \cite{sun2019single, zhang2020portrait} try to solve the problem directly on image level with image-to-image translation. Sun \emph{et al.}\cite{sun2019single} use an encoder-decoder structure based on U-Net~\cite{ronneberger2015u} to implicitly express the unlighted intermediate state and let the network learn the geometry and reflectance information.
Zhang \emph{et al.}\cite{zhang2020portrait} use two GridNet~\cite{fourure2017residual} to remove shadows introduced by external objects, and then to soften shading as well as shadows projected by facial features. Both methods use Light Stage to collect data as it can produce image sets with single lighting sources and simulates arbitrary lighting environments, yet building such a dataset is expensive and time-consuming. There are some other work trying to construct dataset without Light Stage by reconstructing face models \cite{zhou2019deep} or restoring images of Light Stage through color gradient images \cite{meka2019deep}.

Besides relighting portraits on image level, some work first restore the albedo of the input images along with the corresponding normal or mesh, and then render corresponding images under different illuminations. Shu \emph{et al.}\cite{shu2017neural} propose using GAN to end-to-end infer disentangled intrinsic facial attributes, such as shape, albedo and lighting. SfSNet~\cite{sengupta2018sfsnet} also decomposes the image into shape, reflection and lighting, with a residual block structure to learn the relation between high-frequency variation and physical attributes. 
Nestmeyer \emph{et al.}\cite{nestmeyer2020learning} implement delighting and relighting by restoring albedo of the input image through the diffuse physics-based image generation model with Light Stage data.

Unlike the relighting task that requires users to input lighting parameters, our method automatically predicts the lighting parameters to correct the original lighting of input image. Meanwhile, as the reconstruction and rendering process will lead to unavoidable distortion, we do not use the rendered results with corrected lighting as our final outputs like the second category of methods. Instead, we use it to guide our image-to-image translation network for generating more photo-realistic lighting-enhanced images.

\subsection{Differentiable Renderer}
Differentiable renderer takes shape, texture, lighting parameters, and camera pose parameters as input and outputs rendered image and depth image. The most important feature of differentiable renderers is that it makes it possible to calculate the gradient of the rendered image to the input variables. The key challenge is how to make the rasterization process differentiable. To solve this problem, \cite{loper2014opendr,kato2018neural,genova2018unsupervised} try to approximate the gradients in an inverse manner, while \cite{rhodin2015versatile,liu2019soft,chen2019learning} try to simulate the forward rasterization process.
Differentiable renderers are widely used in various applications, including object reconstruction \cite{yan2016perspective, tulsiani2017multi}, human pose estimation \cite{pavlakos2018learning}, light source estimation \cite{nieto2018robust}, and so on. In lighting correction stage of the proposed method, we integrate the differentiable renderer after the lighting parameter correction network to enable applying image-level loss to 3D mesh and lighting parameters. It serves as a bridge to connect the deep neural network and the reconstructed 3D information.


\subsection{Transformer}
Transformer is first proposed in natural
language processing~(NLP)~\cite{vaswani2017attention}, whose success makes it gradually popular in computer vision areas. For example, DETR~\cite{carion2020end} leverages the transformer as the backbone to
cope with the object detection problem. ViT~\cite{dosovitskiy2020image} first applies transformer for image recognition. Wan~\emph{et al.}~\cite{wan2021high} adopt transformer to address high-fidelity free-form pluralistic image completion and have achieved visually pleasing results.
In these works, transformer architecture shows promising performance because of its strong capability to model global image structures by building dense correlations.
However, these works only consider modeling self correlations involved in the input features. In contrast, our method aims to model dense correlations between the input and a guidance image to exploit the texture and light information involved in the guidance to help enhance the input image.

\section{Method}
\begin{figure*}[tbp]
  \centering
  \hfill
  \includegraphics[width=\linewidth]{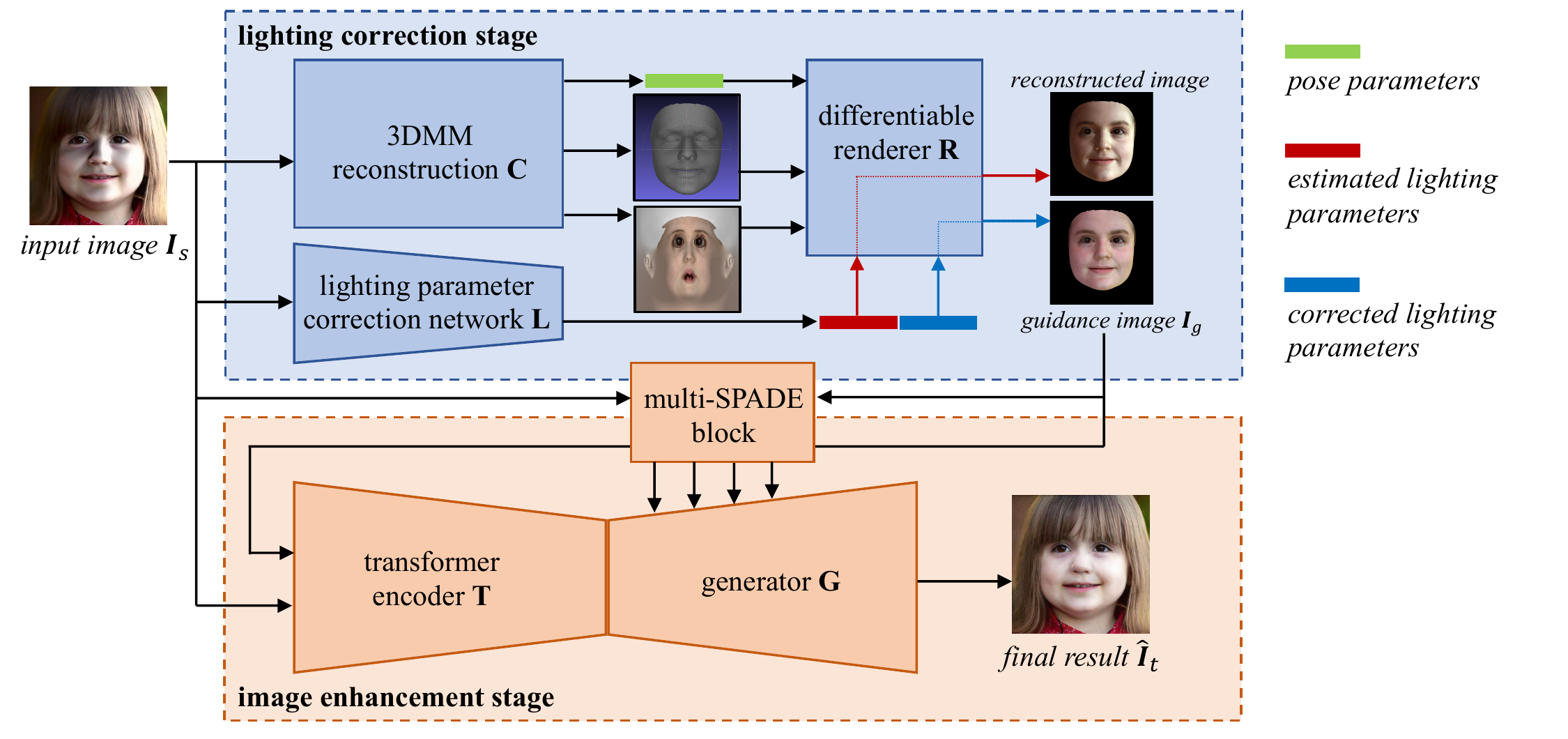}
  \caption{The schematic illustration of our method.}
  \label{fig:framework}
\end{figure*}

Our method aims to enhance the lighting of input portrait images taken under some unsatisfactory lighting conditions. Considering that the shape of human faces is rich in geometric details and the shading on the face is also delicate, we do not follow previous methods which model this task as an image-to-image translation problem and directly learn a mapping from the bad lighting image distribution $P(\mathbf{I}_{s})$ to target light-enhanced image distribution $P(\mathbf{I}_{t})$, defined as $F:P(\mathbf{I}_{s})\rightarrow P(\mathbf{I}_{t})$. Instead, we achieve it in two stages to incorporate 3D information in learning. Specifically, we first learn a mapping from $P(\mathbf{I}_{s})$ to the distribution of the 3D guidance $\mathbf{I}_{g}$, defined as $F:P(\mathbf{I}_{s})\rightarrow P(\mathbf{I}_{g})$, and then we learn a mapping from $P(\mathbf{I}_{s})$ to $P(\mathbf{I}_{g})$ conditioned on the 3D guidance, defined as $P(\mathbf{I}_{s}|\mathbf{I}_{g})\rightarrow P(\mathbf{I}_{t})$. 

Figure \ref{fig:framework} shows our framework. Given an input image $I_s$, in the lighting correction stage~(Section \ref{sec:relighting}), its 3D mesh and albedo map are first reconstructed by a 3DMM reconstruction block $\mathbf{C}$. And then the lighting parameter correction network $\mathbf{L}$ combined with a differentiable renderer $\mathbf{R}$, is learned to predict the corrected lighting parameters, and yields a guidance image $\mathbf{I}_{g}$ containing a face rendered with corrected lighting, for the next stage. The image enhancement stage~(Section \ref{sec:retouching}) is based on a GAN network that consists of a transformer encoder $\mathbf{T}$, a generator $\mathbf{G}$, and a discriminator $\mathbf{D}$. It takes the guidance $\mathbf{I}_{g}$  as a condition to modify the shading of the input face image $I_s$ and eventually generate an enhanced result image, which is visually pleasing and photorealistic with good lighting.


\subsection{Lighting Correction Stage}
\label{sec:relighting}
To include 3D facial information into our framework, we first reconstruct the shape of the input face and disentangle the albedo information from the texture. {Following the idea of \cite{fried2016perspective} which matches 3D models to portraits}, we adopt a reconstruction network from \cite{deng2019accurate}, which takes the input image and use a neural network to regress identity coefficients $\boldsymbol{\alpha}$, expression coefficients $\boldsymbol{\beta}$, and texture coefficients $\boldsymbol{\zeta}$ of a 3DMM face model \cite{paysan20093d} as well as the pose parameters $\boldsymbol{\gamma}$ corresponding to the input. The face shape $\mathbf{S}$ and the albedo texture $\mathbf{A}$ can be presented as:
\begin{equation}
\begin{aligned} \mathbf{S} &=\mathbf{S}(\boldsymbol{\alpha}, \boldsymbol{\beta})=\overline{\mathbf{S}}+\mathbf{B}_{i d} \boldsymbol{\alpha}+\mathbf{B}_{e x p} \boldsymbol{\beta} \\ \mathbf{A} &=\mathbf{A}(\boldsymbol{\zeta})=\overline{\mathbf{A}}+\mathbf{B}_{t} \boldsymbol{\zeta} \end{aligned}
\end{equation}
where \(\overline{\mathbf{S}}\) and \(\overline{\mathbf{A}}\) are the average face shape and texture, \(\mathbf{B}_{i d}\),
\(\mathbf{B}_{e x p}\), and \(\mathbf{B}_{t}\) are the PCA bases of identity, expression, and
texture, respectively. As shading is not considered during the formation of the texture bases, the interpolation result of the texture can only represent albedo information regardless of the input illumination, resulting in a desired disentangle effect.

After reconstruction, we train a lighting correction network to predict lighting parameters corresponding to an enhanced lighting condition.
{To back-propagate gradients from loss functions defined on 2D images to the network, we use a differentiable renderer \cite{liu2019soft} to render the shape and albedo texture of the input image with lighting parameters obtained from the network. The renderer $\mathbf{R}$ replaces the traditional rendering steps of rasterization and hidden face removal with a differentiable aggregate function. For $i^{th}$ pixel, its color can be represented as:
$$
I^{i}=\sum_{j } w_{j}^{i} C_{j}+w_{b}^{i} C_{b},
$$
where $C_{j}$ is the color of $j^{th}$ triangle, $C_{b}$ is the color of the background and $w^{i}$ are weights corresponding to the $i^{th}$ pixel. The weights are negatively correlated with the distance between the pixel and the triangle, as well as the depth of the triangle.}

{For network $\mathbf{L}$, a good prediction requires accurately estimating the input lighting and then mapping the input lighting to its correction target. Experiments show that training the network to learn the two steps simultaneously and to directly output the corrected lighting is difficult. Therefore, we design a bi-branch pipeline to explicitly learn the estimation and the mapping with supervision from input and target images, respectively.}
With the bi-branch design, the lighting parameter correction network encodes two sets of parameters, the estimated lighting parameters $\epsilon_{est}$ and the difference between the estimated lighting parameters and the corrected lighting parameters $\boldsymbol{\delta_{SH}}$, with which the corrected lighting parameters $\epsilon_{crt}$ can be calculated by $\epsilon_{crt}=\epsilon_{est}+\boldsymbol{\delta_{SH}}$. Both $\epsilon_{est}$ and $\epsilon_{crt}$ are input to the differentiable renderer along with the reconstruction results to obtain the reconstructed image $\hat{\mathbf{I}}_{s}=\mathbf{R}(\mathbf{S}, \mathbf{A}, \boldsymbol\gamma, \epsilon_{est})$ and the guidance image with corrected illumination ${\mathbf{I}}_{g}=\mathbf{R}(\mathbf{S}, \mathbf{A}, \boldsymbol\gamma, \epsilon_{crt})$. The loss function can be written as:
\begin{equation}
\label{func_relit}
\mathcal{L}_{light}=\left|\hat{\mathbf{I}}_{s}-\mathbf{I}_{s}\right|+\lambda_{crt}\left|{\mathbf{I}}_{g}-{\mathbf{I}}_{t}\right|
\end{equation}
where $\lambda_{crt}$ is a weight parameter.

Although the corrected lighting parameters are obtained and used in the rendering process, the rendered image cannot be directly used as the enhancement result, as there is unavoidable distortion in the reconstruction process and the lighting model in the renderer cannot perfectly simulate the real lighting effects. Therefore, we use an image-to-image translation network to guarantee the reality of the enhancement result, while using the rendered image as a guidance to provide the translation process with realistic shading and texture information.

\subsection{Image Enhancement Stage}
\label{sec:retouching}
Given the input low-light face image $\mathbf{I}_s$ and the guidance image $\mathbf{I}_{g}$ generated in the lighting correction stage, the image enhancement stage outputs the desired enhanced face image.
As mentioned in Section \ref{sec:intro}, to better exploit the visual relations between $\mathbf{I}_s$ and $\mathbf{I}_{g}$ is the key to reconstruct the desired enhanced face image. We apply two modifications on the basis of pix2pix \cite{isola2017image}, the transformer encoder and the multi-SPADE block, to make better use of the guidance from the lighting correction stage and generate visually-pleasing results.


\begin{figure}[htb]
   \includegraphics[width=\linewidth]{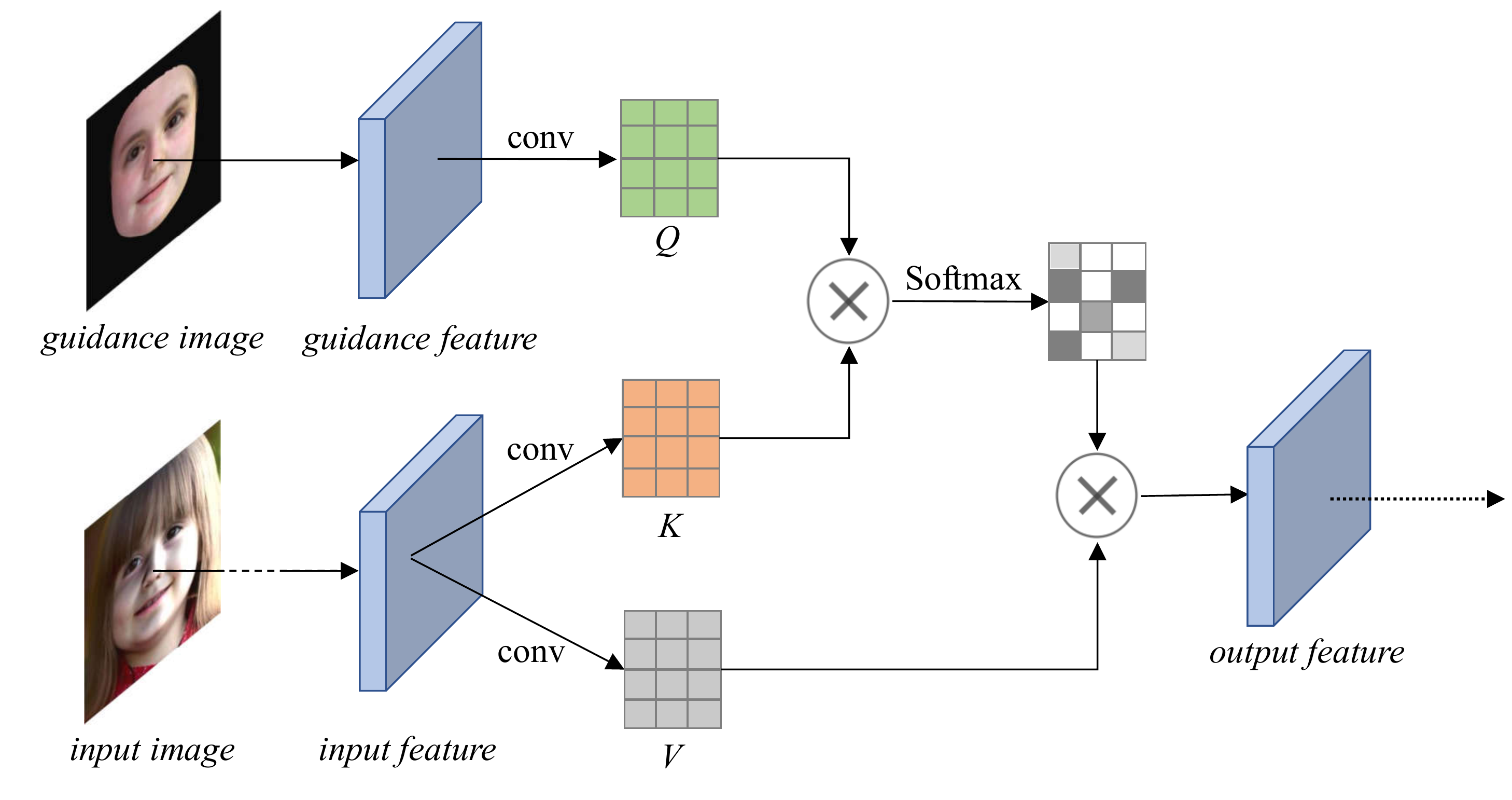}
   \caption{Transformer block.}
    \label{fig:transformer}
\end{figure}

\begin{figure}[htb]
   \includegraphics[width=\linewidth]{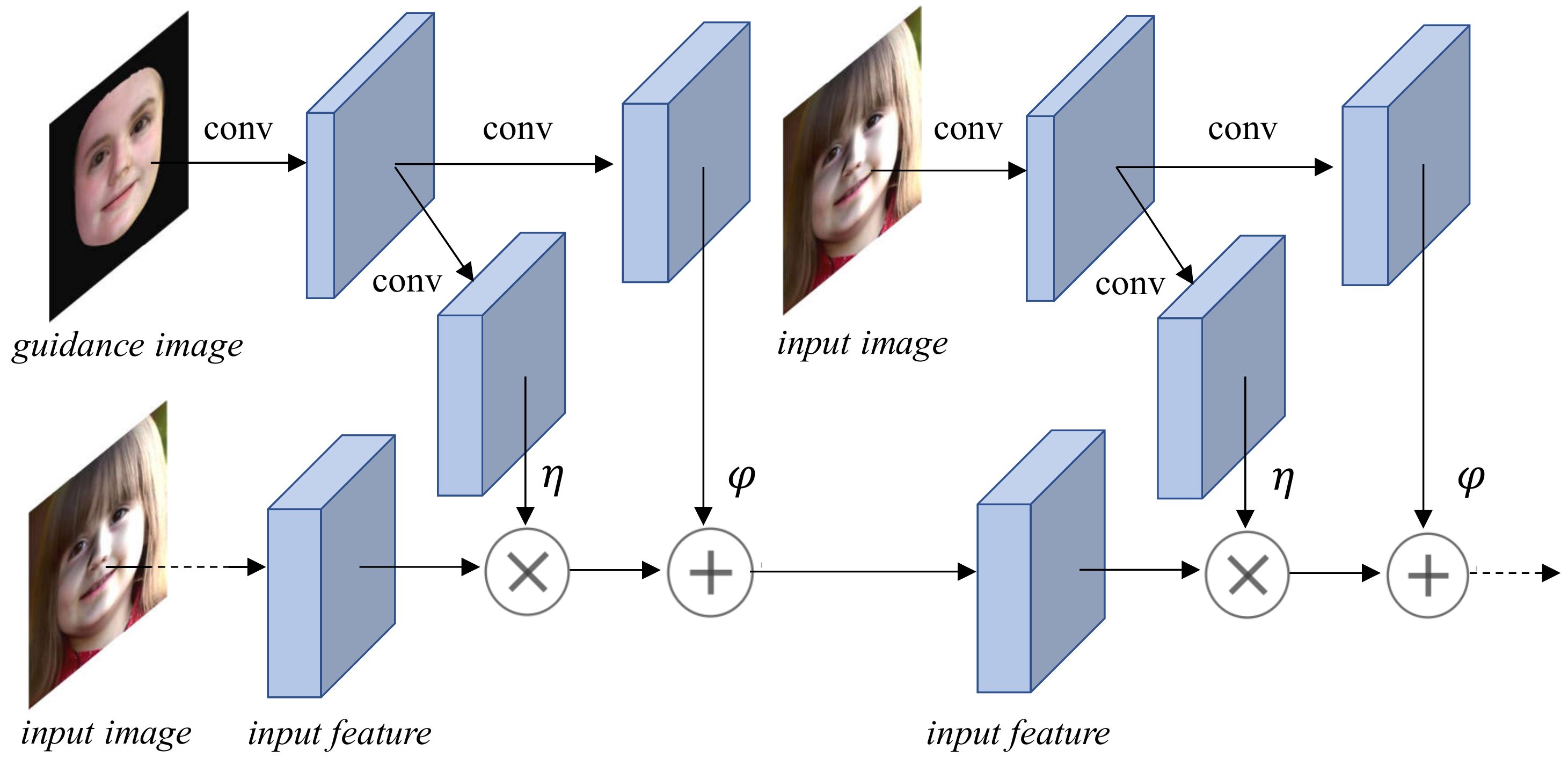}
   \caption{
     Multi-SPADE block. {$\eta$ and $\varphi $ are modulation parameters.}}
    \label{fig:multi_spade}
\end{figure}

\textbf{Transformer Encoder}:
Traditional transformers capture self correlations by calculating $Q$~(Query), $K$~(Key), and $V$~(Value) using the input feature, but we want to model the relations between the input and the guidance. Therefore, we propose a transformer block as shown in Figure~\ref{fig:transformer}.
In the proposed block, the input feature is extracted from the input image and the guidance $\mathbf{I}_g$ will first be warped into the same dimension as the input feature. Then $Q$ is calculated from $Warp(\mathbf{I}_g)$, while $K$ and $V$ are calculated from the input feature. The attention map is obtained by performing a matrix multiplication between $Q$ and $K$ and applying a softmax operation to its result.
By multiplying $V$ and the attention map, we acquire the output feature.

\textbf{Multi-SPADE Block}: 
SPADE~(Spatially-Adaptive Normalization) \cite{park2019semantic} is an
effective layer in a generator for synthesizing photo-realistic images
given an input semantic layout. Unlike the original SPADE block  which only injects the guidance image as condition and totally ignores the semantics involved in the original input image, 
we adopt a Multi-SPADE block from \cite{mallya2020world} to make the generated image better preserve semantics-consistence with the input. Specifically, we cascade two SPADE blocks, one takes $\mathbf{I}_{g}$ as input and the other one takes $\mathbf{I}_s$ as input, as shown in Figure~\ref{fig:multi_spade}. Such a cascaded block ensures the generated image be consistent with $\mathbf{I}_{g}$ at light level and simultaneously be semantics-consistent to the input $\mathbf{I}_s$. 

There are three loss functions used in training the enhancement network. We use the adversarial loss to minimize the distribution distance between the ground-truth and output normal light distributions. For a face image, some local areas require much attention to adaptive but a global discriminator fail to provide the desired adaptivity. Similar to PatchGAN \cite{isola2017image}, we use a multi-scale patch discriminator $\mathbf{D}=\left \{  \mathbf{D}_1,\mathbf{D}_2,\mathbf{D}_3\right \}$ to discriminate the real and fake images at different scales. Thus, the adversarial loss is defined as follows:
\begin{equation}
\label{func_gan_obj}
\min _{T,G}\max _{D_{1}, D_{2}, D_{3}} \sum_{k=1,2,3} \mathcal{L}_{\mathrm{GAN}}\left(T, G, D_{k}\right)
\end{equation}
\begin{equation}
\label{func_gan}
\mathcal{L}_{\mathrm{GAN}}=\mathbb{E}_{\mathbf{I}_{t}}[\log D_k(\mathbf{I}_{t})]+\mathbb{E}_{(\mathbf{I}_s, \mathbf{I}_g)}[\log (1-D_k(G(T(\mathbf{I}_s),\mathbf{I}_g)))]
\end{equation}

To make the training process robust, we also adopt a feature matching loss \cite{wang2018high} between different layer features extracted by the discriminator of the real and fake images. 
\begin{equation}
\label{func_FM}
\mathcal{L}_{\mathrm{FM}}\left(T,G, D_{k}\right)=\mathbb{E}_{(\mathbf{I}_s, \mathbf{I}_g,\mathbf{I}_t)} \sum_{i=1}^{L} \frac{1}{N_{i}}\left[\left\|D_{k}^{(i)}(G(T(\mathbf{I}_s),\mathbf{I}_g))-D_{k}^{(i)}(\mathbf{I}_{t}))\right\|_{1}\right]
\end{equation}
where $i$ means the $i^{th}$ layer, $L$ represents the number of layers, and $N_i$ is the number of parameters of the $i^{th}$ layer.

Also, a perception loss is used to further improve the performance:
\begin{equation}
\label{func_percep}
\mathcal{L}_{percep}(T,G)= \sum_{i=1}^{K} \frac{1}{M_{i}}\left[\left\|F^{(i)}(\mathbf{I}_s)-F^{(i)}(G(T(\mathbf{I}_s),\mathbf{I}_g)))\right\|_{1}\right]
\end{equation}
where $i$ means the $i^{th}$ layer, $K$ represents the number of layers, $M_i$ is the number of parameters of the $i^{th}$ layer, and $F^{(i)}$ is the feature of the $i^{th}$ layer of a VGG19 network.

In our work, these losses can be optimized jointly and the total objective can be defined as:
\begin{equation}
\begin{split}
\label{func_gan_obj}
\min _{T,G}\max _{D_{1}, D_{2}, D_{3}} \sum_{k=1,2,3} \mathcal{L}_{\mathrm{GAN}}\left(T,G, D_{k}\right)+ \\
\lambda_{FM} \sum_{k=1,2,3} \mathcal{L}_{\mathrm{FM}}\left(T,G, D_{k}\right)+ 
\lambda_{percep} \mathcal{L}_{percep}\left(T,G\right)
\end{split}
\end{equation}
where $\lambda_{FM}$ and $\lambda_{percep}$ are weight coefficients.

\section{Experiments}
\subsection{Dataset Construction}

\begin{figure}[htb]
   \includegraphics[width=\linewidth]{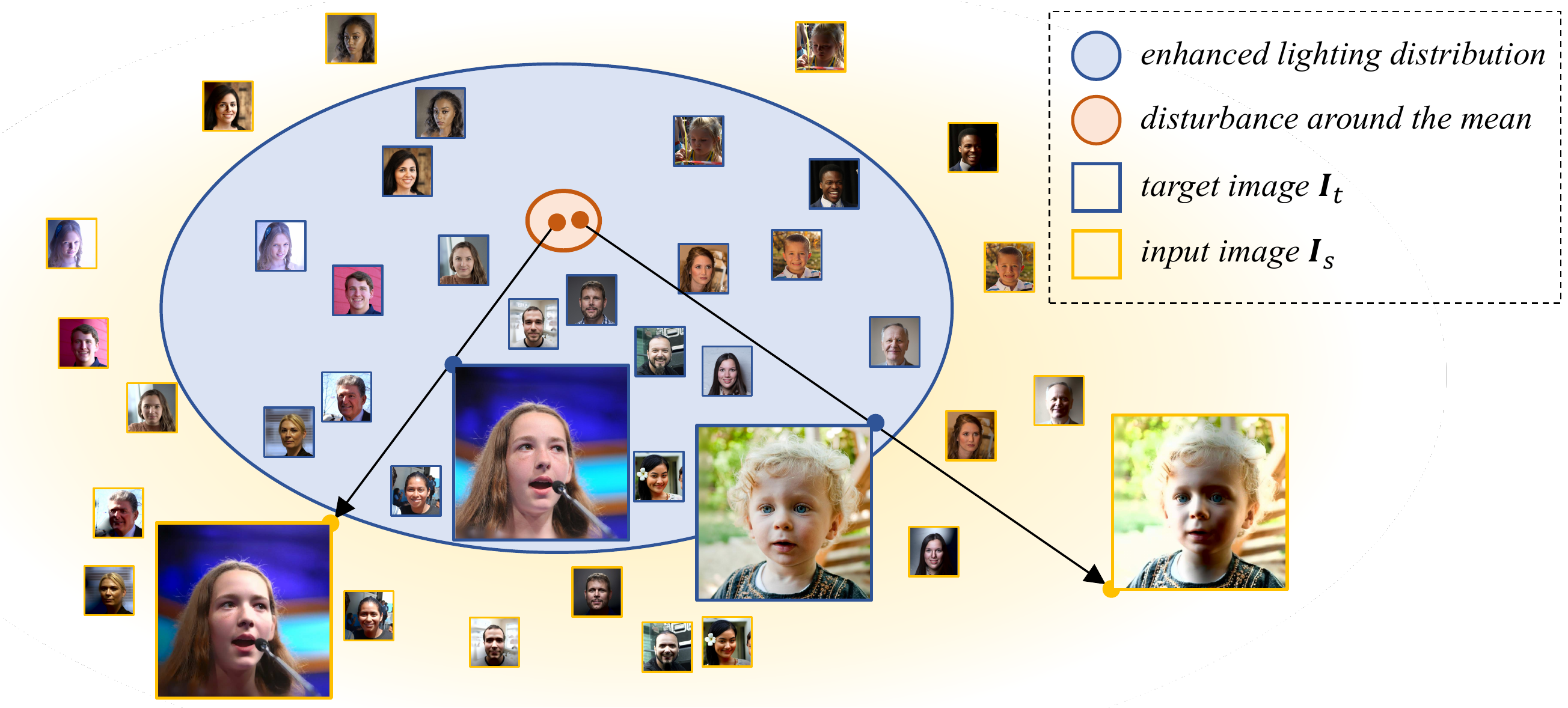}
   \centering
   (a) Illustration of data synthesis \\
   \includegraphics[width=\linewidth]{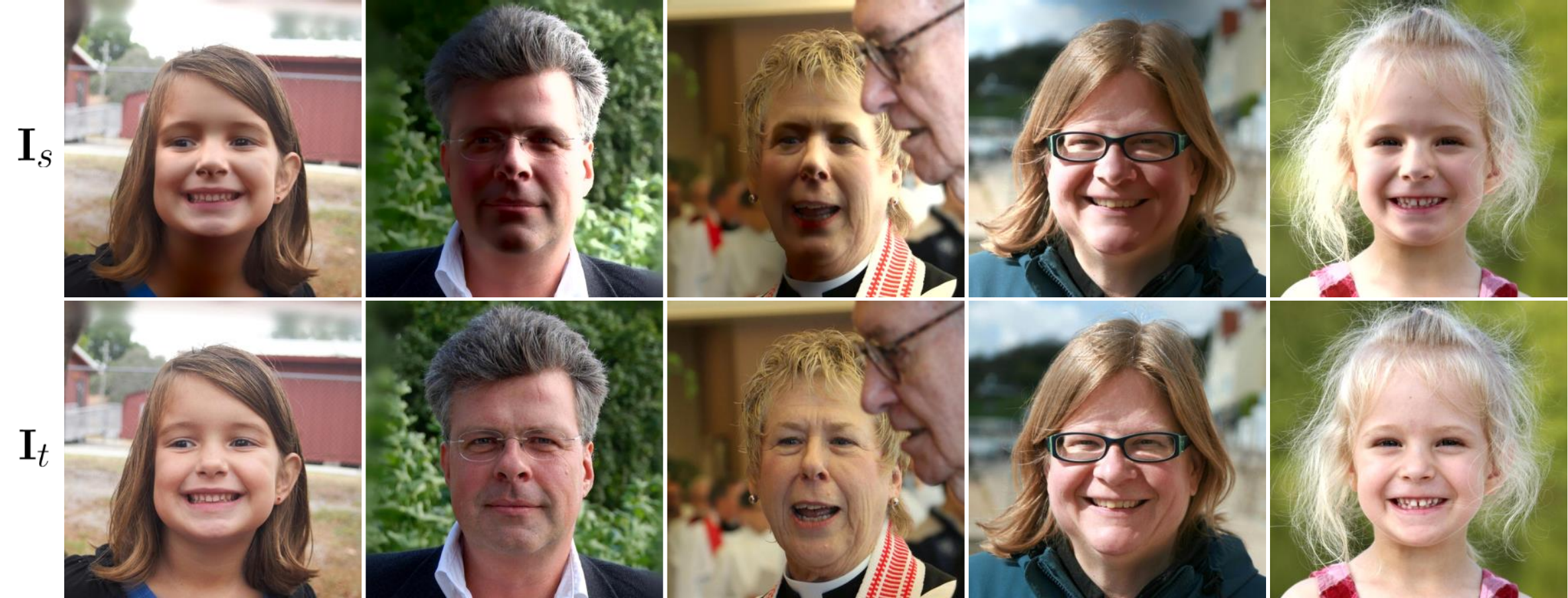}
   (b) Samples of synthesized training pairs. \\
   \caption{Dataset construction process and samples.
      $\mathbf{I}_s$ is the degraded low-light image and $\mathbf{I}_{t}$ is the pseudo ground-truth image.}
    \label{fig:sample}
\end{figure}

As mentioned in Section \ref{sec:intro}, currently there exists no high-quality dataset for portrait lighting enhancement. To build such a dataset, we first ask volunteers to manually select a group of portrait images that are believed to have good lighting from FFHQ dataset~\cite{karras2019style}. This group of images forms the enhanced lighting distribution as shown by the blue oval of Figure~\ref{fig:sample}. They will be used as the ground truth for training to guarantee the outputs to be more natural than using synthetic data. The corresponding input images are degraded from target images. Specifically, we first estimate SH parameters of all target images following the iterative optimization method used in \cite{zhou2019deep} and calculate their mean value $\overline{SH}$. Given the SH parameters of $i^{th}$ target image, denoted as $SH_{target,i}$, we randomly sample an ideal SH parameter within a small radius of the mean, denoted as $\overline{SH}_i$, and extrapolate $SH_{target,i}$ away from $\overline{SH}_i$ to get the degraded lighting: $SH_{input, i} = SH_{target, i} + \lambda_{sh} * (SH_{target, i} - \overline{SH}_i) $ where {$SH_{input, i}$ represents the SH parameters of the $i^{th}$ input image} and $\lambda_{sh} \in [1,2]$ is a randomly sampled extrapolation parameter. 
The randomness in ideal point and extrapolation parameter sampling introduce lighting variations in dataset. With degraded lighting parameters $SH_{input, i}$, the $i^{th}$ degraded image is generated by the data synthesis pipeline introduced in \cite{zhou2019deep}. {After degradation, to make the synthetic data as realistic as possible, a blurring filter is applied to the degraded image, followed by a Gaussian noise to simulate detail loss and noise artifacts in poorly-relit images. The strength of the blurring filter and the Gaussian noise is not uniform but in positive correlation with the shading darkness.} And thus, the training pair of the $i^{th}$ input and target images is formed.

{To further prevent unrealistic synthetic images, volunteers are asked to perform manual filtering, after which 6828 out of 20000 image pairs are selected to form the final dataset.
Among the 6828 pairs, 6371 pairs are splited as the training set, while 457 pairs are splited as the testing set.} 
Some samples from the dataset are shown in Figure \ref{fig:sample}. 

{This dataset construction method is pecuniarily efficient and the dataset is easy to extend. Training with the dataset, networks will learn to adjust the lighting of the input images towards the elected standard, yet the results will still preserve the lighting characteristic of the input images, instead of undesirably mapping all lighting conditions to a normalized one. Although the domain gap between synthetic data and real data is inevitable, even with the above-mentioned delicate designs, the testing results on real images demonstrate that our synthetic dataset enables networks to generalize well on real poorly-relit images.}

\subsection{Implement Details}



In the lighting correction stage, $\mathbf{L}$ is modified from the structure of ResNet50 \cite{he2016deep}, which concatenates the $\mathbf{I}_s \in R^{512\times512\times3}$ and $\mathbf{I}_g\in R^{512\times512\times3}$ as input and outputs a matrix of size $B\times 54$, where $B=4$ is the batch size and 54 is the vector length representing 2 groups of $3\times 9$ parameters, $\epsilon_{est}$ and $\boldsymbol{\delta_{SH}}$.
{The numbers of SH we used are 27 (9 for each channel of RGB). }.
During the training of the lighting correction stage, we optimize the network parameters using Adam with $\beta_1=0.95$ and $\beta_2=0.90$. The initial learning rate is 0.001 and linearly decays for every 15 epochs by a factor of 0.7, with total epochs being 120.
The value of $\lambda_{crt}$ is set to 1.

In the image enhancement stage, $\mathbf{E}$ has 7 convolution layers and 4 transformer blocks in total. It takes $\mathbf{I}_s \in R^{512\times512\times3}$ and $\mathbf{I}_g\in R^{512\times512\times3}$ as input and outputs a feature map of shape $B\times 1024\times 8\times 8$, where $B=16$ is the batch size. 
$\mathbf{G}$ has 7 upsampling layers with scale factor 2, and each is accompanied by a Multi-SPADE block. It reconstructs an image of the same size as the input. 
$\mathbf{D}$ has a similar architecture as PatchGAN\cite{isola2017image} and downsampling factors for $\mathbf{D}_1, \mathbf{D}_2, \mathbf{D}_3$ are 1, 2, and 4 respectively.
We optimize the network parameters using Adam 
with $\beta_1=0.95$ and $\beta_2=0.90$. The initial learning rate is 0.0002 and linearly decays to 0 after 50 epochs.
Values of $\lambda_{FM}$ and $\lambda_{percep}$ are all set as 10. {Besides, in Equation~\ref{func_FM}, $L$ equals 5 including 1 input layer, 2 intermediate layers, and 1 output layer. And in Equation~\ref{func_percep}, $K$ equals 5 which means we apply the perceptual loss to the output of the first 5 layers of the VGG19 network.}

\begin{figure*}[htb]
   \includegraphics[width=\linewidth]{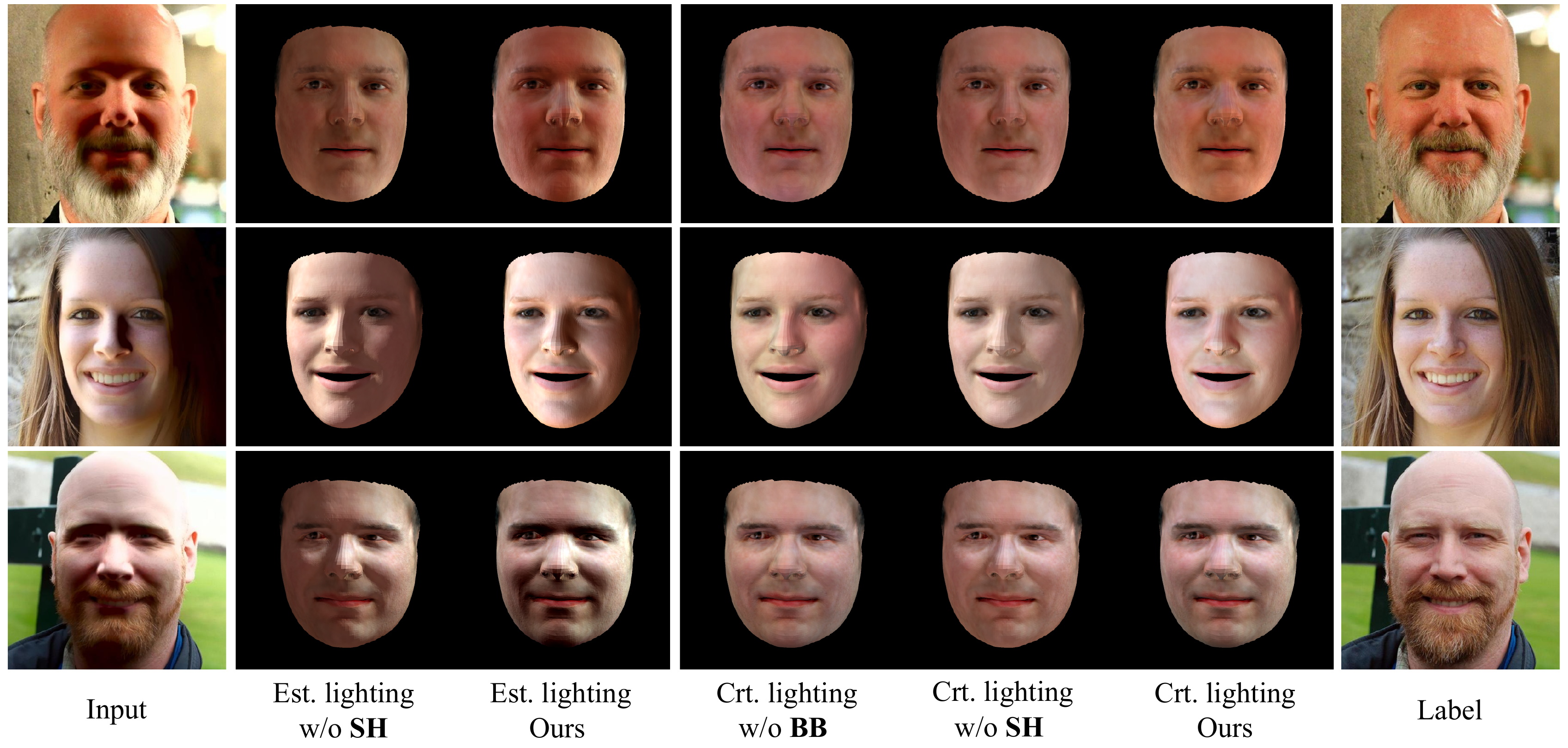}
   \caption{Example results of ablation study on the lighting correction stage.}
    \label{fig:relighting}
\end{figure*}

\subsection{Ablation Study}

\begin{table}[htb]
\centering
\tabcolsep=4pt
\begin{tabular}{c|cccccc}
\toprule
 & Albedo & w/o $\mathbf{BB}$ & w/o $\mathbf{SH}$ & Ours\\ \midrule
PSNR  & 9.868 & 9.994 & 10.17 & \textbf{10.25}\\ \midrule
SSIM  & 0.2774 & 0.2798 & 0.2813 & \textbf{0.2842}\\
\bottomrule
\end{tabular}
\caption{Quantitative evaluation results of ablation study on the lighting correction stage.}
\label{tab:ablation_relighting}
\end{table}

\textbf{Lighting correction stage}

To demonstrate the effectiveness of the bi-branch design and the SH lighting model of the lighting correction stage, we conduct an ablation experiment including two comparisons with three results:
\begin{itemize}
\item Est. lighting w/o $\mathbf{SH}$: the estimated lighting of the method using the default ambient and directional lighting model instead of SH lighting model (\textbf{SH}) in the differentiable renderer. 
\item Crt. lighting w/o $\mathbf{BB}$: the corrected lighting of the method not using the bi-branch design (\textbf{BB}) but use a single-branch to directly estimate the corrected lighting. 
\item Crt. lighting w/o $\mathbf{SH}$: the corrected lighting of the method using the default ambient and directional lighting model in the differentiable renderer. 
\end{itemize}

Some example results are given in Figure \ref{fig:relighting}.
To only output the corrected lighting parameters under the supervision of the ground truth is the most straightforward design of the lighting parameter
correction network. However, in such a single-branch design without explicitly estimating the original lighting of the input image, it will cause ambiguities and lead to results sometimes being too close to the input (2nd example) but sometimes being too flat and losing the original lighting patterns (1st and 3rd examples). In our bi-branch design, the network can learn to estimate and adjust the lighting from separate supervision, which leads to better performance.

Another issue that influences the lighting correction performance is that the original differentiable renderer supports only ambient lighting model and direction lighting model, which sometimes cannot perfectly simulate the lighting effect of the image. In our implementation, we modify the lighting model to SH lighting, which is more flexible in simulating lighting directions and colors. In Figure \ref{fig:relighting} we can see that the SH lighting model is able to simulate more complex lighting condition and better fit the skin color of the input image, therefore leads to better corrected lighting results.

We also conduct quantitative evaluation by calculating Peak Signal-to-Noise Ratio~(PSNR) and Structural Similarity~(SSIM) against target images. The results in Table~\ref{tab:ablation_relighting} shows that our final method obtains the highest score on both metrics, which numerically proves the effectiveness of the bi-branch design with SH lighting model.


\begin{table}[htb]
\centering
\tabcolsep=4pt
\begin{tabular}{c|cccccc}
\toprule
 & Input & w/o $\mathbf{I}_g$ & w/o $\mathbf{SH}$ & w/o $\mathbf{T}$ &  w/o $\mathbf{MS}$ & \textbf{Ours}\\ \midrule
PSNR  & 18.25 & 24.21  & 26.13 & 26.26 & 26.93& \textbf{27.15}\\ \midrule
SSIM  & 0.8816 & 0.9030  & 0.9115 & 0.9163 & 0.9188 & \textbf{0.9211}\\
\bottomrule
\end{tabular}
\caption{Ablation study on the image enhancement stage.}
\label{tab:ablation_study}
\end{table}

\textbf{Image enhancement stage}

To demonstrate the effectiveness of each part of the image enhancement stage, we conduct an ablation experiment to compare our method with four methods:
\begin{itemize}
\item w/o $\mathbf{I}_g$: the method that removes the 3D guidance injection $\mathbf{I}_g$ in the retouching branch. 
{The input of this method is the image $\mathbf{I}_s$ only. Specifically, we set $\mathbf{I}_s$ as the input to the encoder $\mathbf{T}$ without using Multi-SPADE blocks and the guidance $\mathbf{I}_g$.}
\item w/o SH: the method that uses guidance if point lighting rather than SH lighting in the relighting branch.
\item w/o $\mathbf{T}$: the method that adopts traditional convolution layers as encoder rather than the transformer encoder.
\item w/o $\mathbf{MS}$: the method that uses single SPADE block which only injects $\mathbf{I}_g$ without considering semantics consistency using Multi-SPADE blocks.
\end{itemize}

\begin{figure*}[htb]
  \centering
  \hfill
  \includegraphics[width=\linewidth]{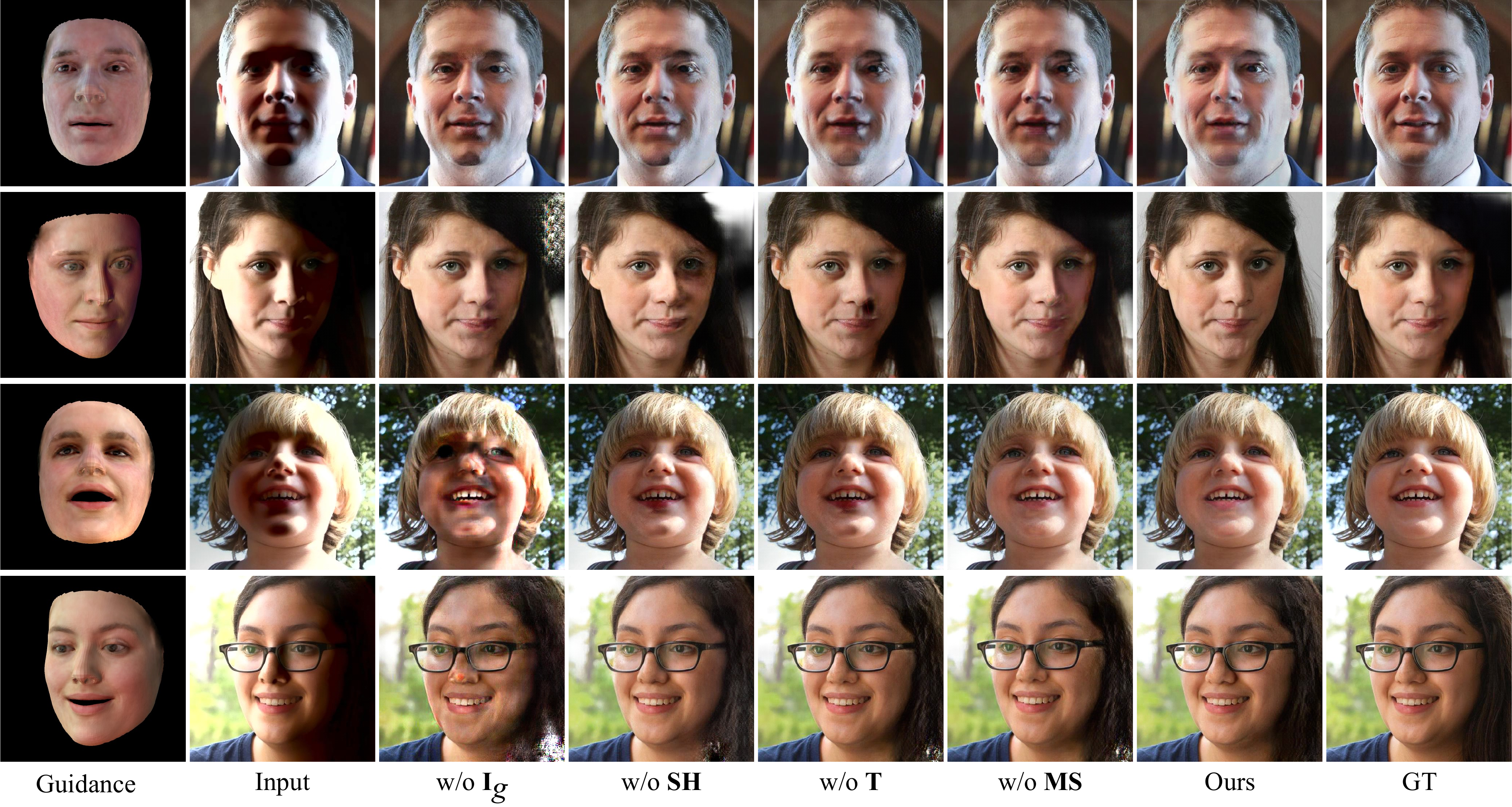}
  \caption{
  Visual results of ablation study on the image enhancement stage.}
  \label{fig:abla_ffhq}
\end{figure*}


First, qualitative experiments are conducted, as shown in  Figure~\ref{fig:abla_ffhq}.
We observe that method w/o $\mathbf{I}_g$ has obtained the worst results of all ablations, which evaluates the necessity of using 3D guidance for image enhancement. 
Sometimes method w/o SH shows enhancements compared to the input, but it is still not comparable to ours. For example, the lower jaw of the 3rd example cannot be enhanced. This is because the point lighting model fails to estimate the lighting accurately and an inaccurate guidance cannot provide a right training direction of the model. Method w/o $\mathbf{T}$ shows severe artifacts, for example the nose of the girl in the second row. This shows that the low-light regions cannot be well-enhanced without the transformer's ability of modeling correlations between the input and the guidance. 
Method w/o $\mathbf{MS}$ outperforms other ablations, but the absence of texture details makes its results non-photorealistic. For example, we cannot see enough details of the 3rd and 4th examples.
Depending only on single SPADE block fails to preserve enough texture details.
In contrast,
our method gives superior results, which demonstrates the superior performance and generalization ability of each component of our method.

We also conduct quantitative experiments in Table~\ref{tab:ablation_study}. In contrast, the PSNR and SSIM metrics of our final method are all better than other settings, which validates the effectiveness of our method by imposing 3D guidance, SH lighting model, transformer encoder, and Multi-SPADE block. We also find more serious degradation
without using our proposed 3D guidance than others. This reveals the importance of providing 3D information to the enhancement process. This 3D guidance not only guides the training process in the right direction, but also provides missing texture details of the input, especially in the underexposure and overexposure regions. 
Though the 3D guidance contributes more to the improvement of our method, other components are also crucial to the success of the final pleasing results.

\subsection{Comparisons on Synthetic Data}
To evaluate the performance of the proposed method, we compare it with state-of-the-art image lighting enhancement methods, EnlightenGAN~\cite{jiang2019enlightengan} and Zero-DCE~\cite{guo2020zero}, and portrait relighting methods, DSIPR~(Deep Single-Image Portrait Relighting)~\cite{zhou2019deep} and SMFR~\cite{hou2021towards}. We train the two image lighting enhancement methods on our dataset with author-released code, as they were previously trained for general images, and use the pre-trained model for the portrait relighting methods. As the portrait relighting methods require lighting parameters as input, we feed them with the lighting parameters predicted by our first stage.

\begin{figure*}[htb]
  \centering
  \hfill
  \includegraphics[width=\linewidth]{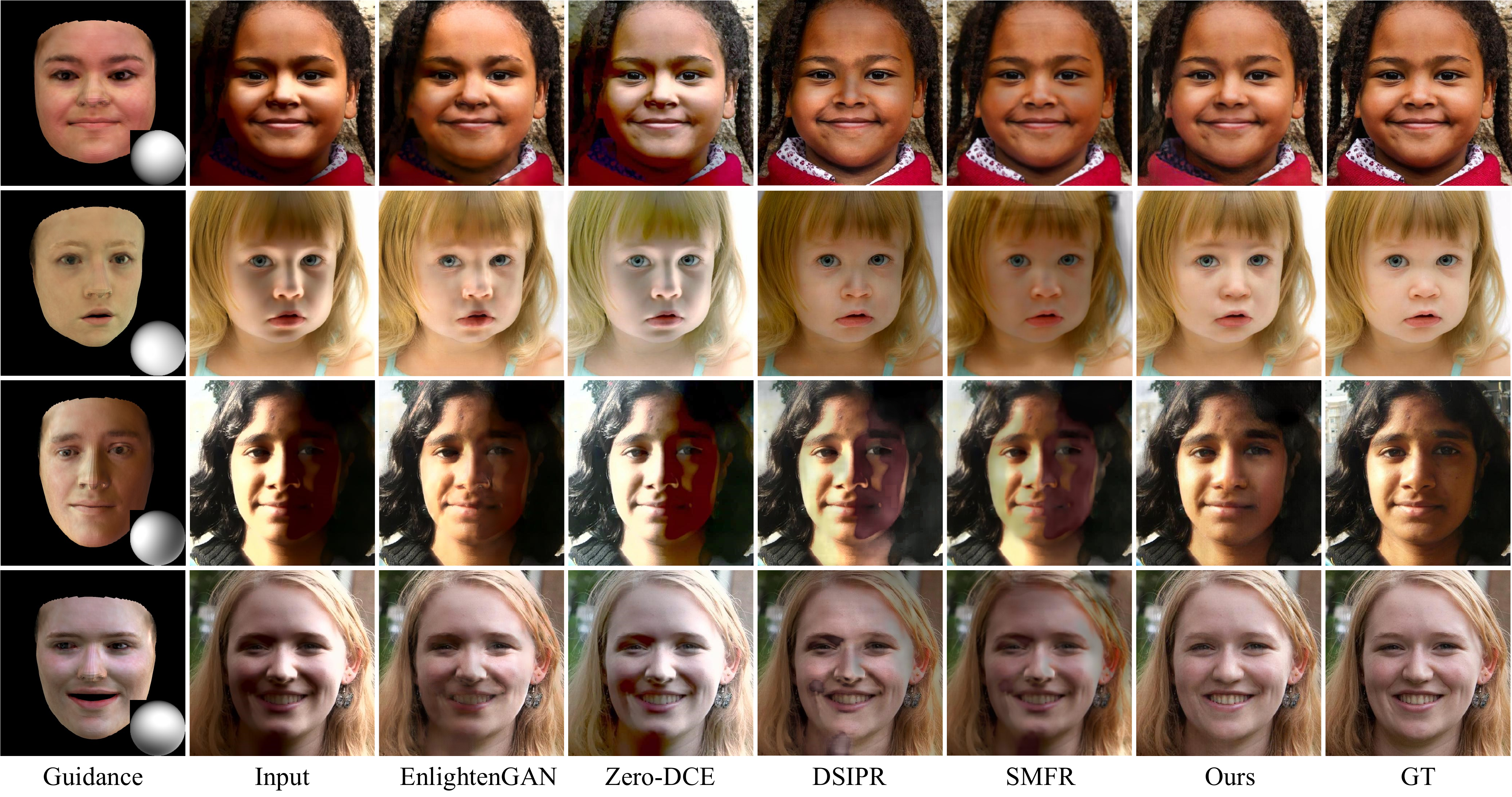}
  \caption{Qualitative comparison with state-of-the-art methods on the FFHQ dataset on the image enhancement stage.}
  \label{fig:comp_ffhq}
\end{figure*}

\begin{figure*}[htb]
  \centering
  \hfill
  \includegraphics[width=\linewidth]{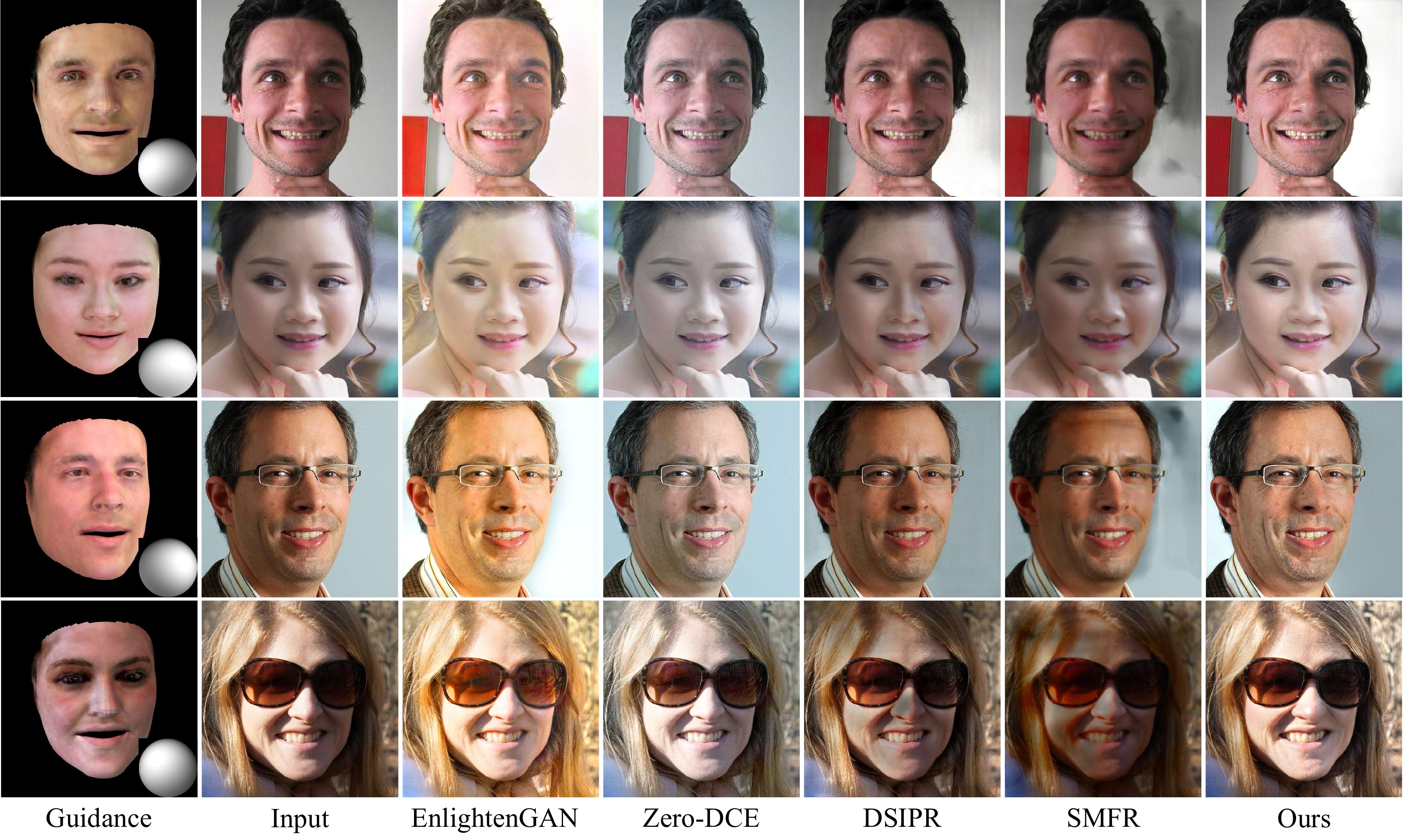}
  \caption{Qualitative comparison with state-of-the-art methods on in-the-wild images on the image enhancement stage.}
  \label{fig:comp_real}
\end{figure*}

We first compare these methods on synthetic data, both qualitatively and quantitatively, results of which are shown in Figure~\ref{fig:comp_ffhq} and Table~\ref{tab:compare}.
In terms of very dark areas in Figure~\ref{fig:comp_ffhq}, all of the comparison methods cannot produce desired results. EnlightenGAN and Zero-DCE achieve better results than the relighting methods DSIPR and SMFR, but they still perform badly in generating natural face textures and pleasing lighting conditions compared to ours. 
Numerical results in Table~\ref{tab:compare} also support this observation. DSIPR and SMFR report the worst two results on both PSNR and SSIM metrics, while EnlightenGAN and Zero-DCE show relatively better results. However, it is obvious that the proposed method has achieved the best in terms of PSNR and SSIM metrics. This proves the superiority of the proposed method in making use of 3D information and correlations between the 3D guidance and the input.


\begin{table}[htb]
\centering
\tabcolsep=2pt
\begin{tabular}{c|cccccc}
\toprule
 & Input & DSIPR  & SMFR & EnlightenGAN & Zero-DCE &  Ours\\ \midrule
PSNR  & 18.25 & 23.68  & 22.77 & 26.15 & 26.32 & \textbf{27.15}\\ \midrule
SSIM  & 0.8816 & 0.8970 & 0.8893 & 0.9088 & 0.9102 & \textbf{0.9211}\\
\bottomrule
\end{tabular}
\caption{Comparison with state-of-the-art models.}
\label{tab:compare}
\end{table}

\subsection{Comparisons on In-the-Wild Data}

\begin{figure}[t]
\centering
\setlength{\tabcolsep}{0.05mm}{
      \includegraphics[width=0.9\linewidth]{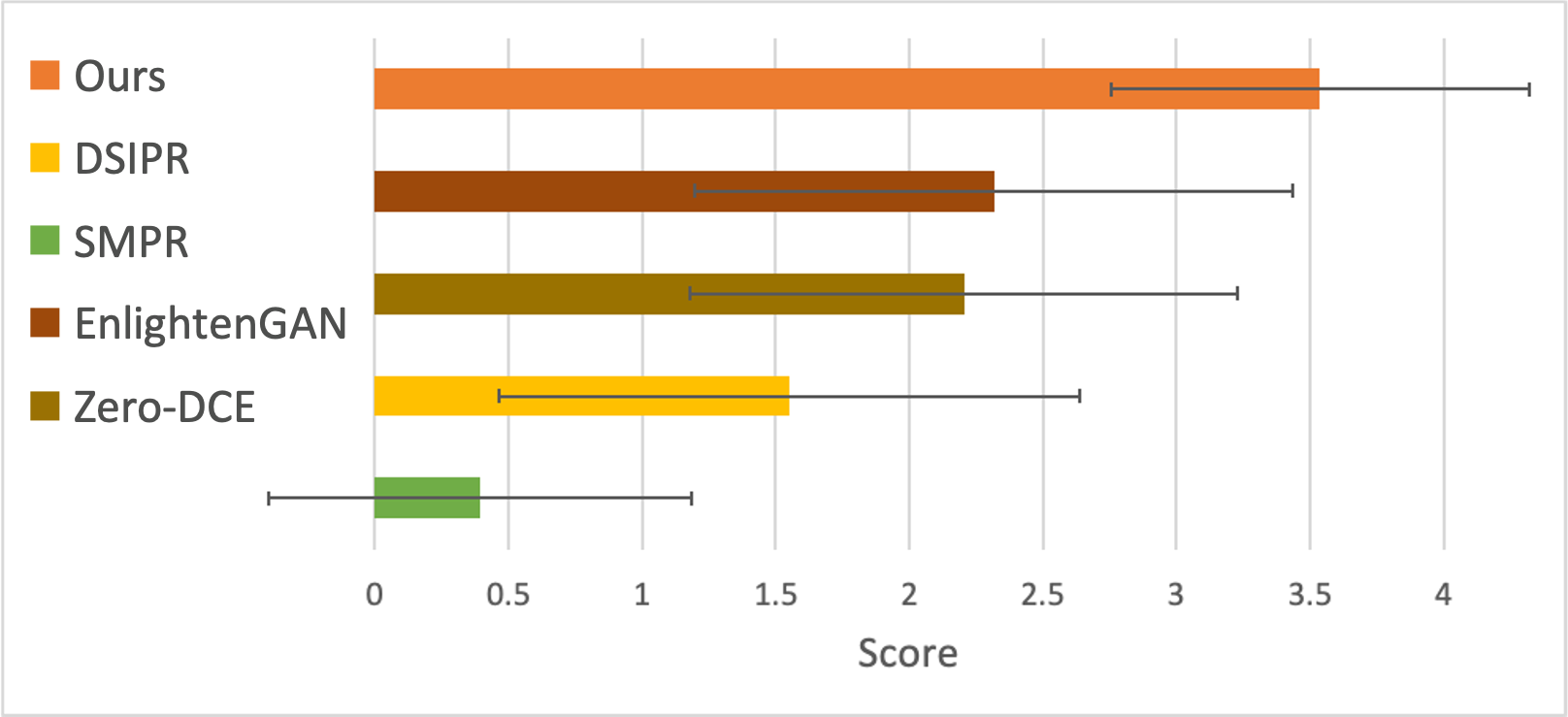} \\
      (a) Average ranking score and standard deviation \\
      \includegraphics[width=0.9\linewidth]{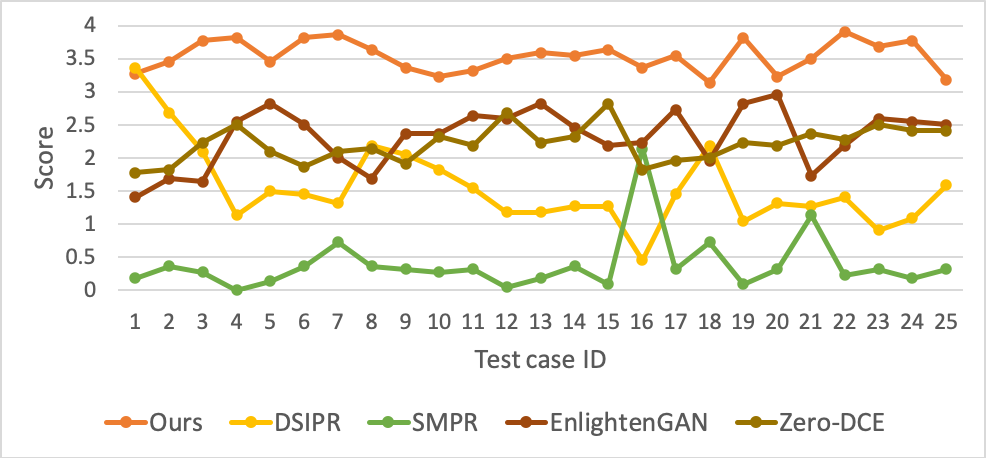} \\
      (b) Ranking score per test case \\
      \includegraphics[width=0.9\linewidth]{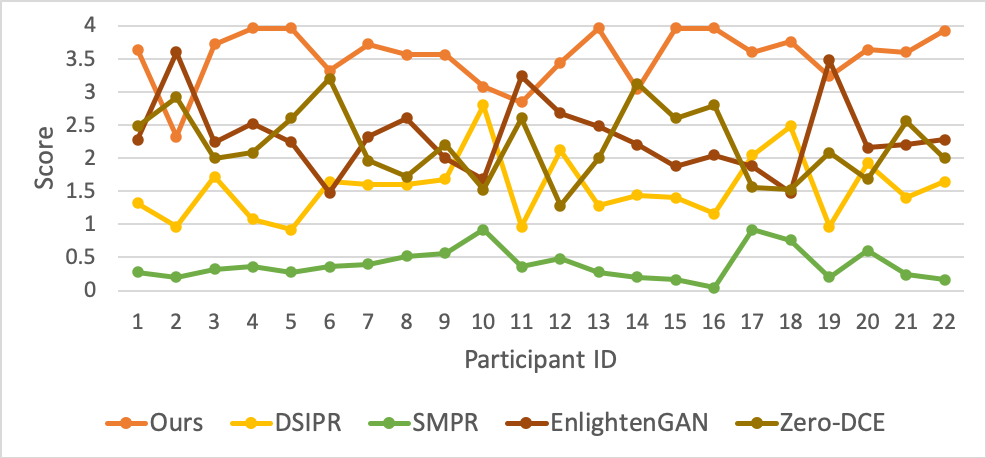} \\
      (c) Ranking score per participant. \\
  }
 \caption{{User study result analysis.} }
 \label{fig:user_study}
\end{figure}

To evaluate the generalization ability of our method, we compare to related works on the in-the-wild images, as shown in Figures \ref{fig:comp_real}.
All methods can enhance the lighting condition overall, but these related works still show obvious worse results than ours. Both EnligtenGAN and Zero-DCE show unrealistic skin colors after enhancing the input. For example, EnligtenGAN makes the skin color more yellow for the face of the 3rd example. Besides, Zero-DCE tends to generate over-exposure and unnatural images. In contrast, our method can successfully enhanced the input and synthesize photo-realistic light-enhanced image without destroying the face color. This is because the 3D information such as face geometry, face texture and lighting directions and intensities involved in the guidance can guide our networks to generate desired enhanced images, while EnligtenGAN and Zero-DCE do not use any guidance.
As for relighting-based methods DSIPR and SMFR, they cannot reconstruct face details. Though they use SH parameters as guidance to train their models, their performances are still worse than ours. This is because these SH parameters only capture lighting information, while the 3D guidance we used cannot only model target lighting but also capture face geometry, shading, and face texture. To conclude, all these methods generate unsatisfactory visual results when it comes to both brightness and naturalness. Compared to these methods, our method successfully enhances the low-right areas and also preserves facial texture details without inducing artifacts. More results please refer to our supplemental materials.


As there is no ground truth for the enhancement of in-the-wild images, we also conduct a user study to perform numerical evaluation among our method and the related works.
Specifically, we randomly choose 25 bad-lighting real images as test cases and apply different methods to enhance these images.
Then for each case, we show the input image and the enhanced images generated by different methods and ask 22 participants to rank the five results from the highest quality (score 4) to the lowest quality (score 0).
During ranking, the participants are instructed to consider whether the lighting condition of the images are properly and naturally enhanced without any artifacts or noise. 
The statistics results are shown in Figure \ref{fig:user_study}. Figure \ref{fig:user_study} (a) demonstrate the average ranking scores and standard deviations of each method, where our method is ranked in the first place (3.54±0.79). Figure \ref{fig:user_study} (b) shows the average ranking scores for each methods by every test case, and Figure \ref{fig:user_study} (c) shows the average ranking scores for each methods by every participant, from both of which it can be seen that our method has a stable and outstanding performance over different test cases judged by different individuals.

\subsection{Image Harmonization}

\begin{figure}[htb]
   \includegraphics[width=\linewidth]{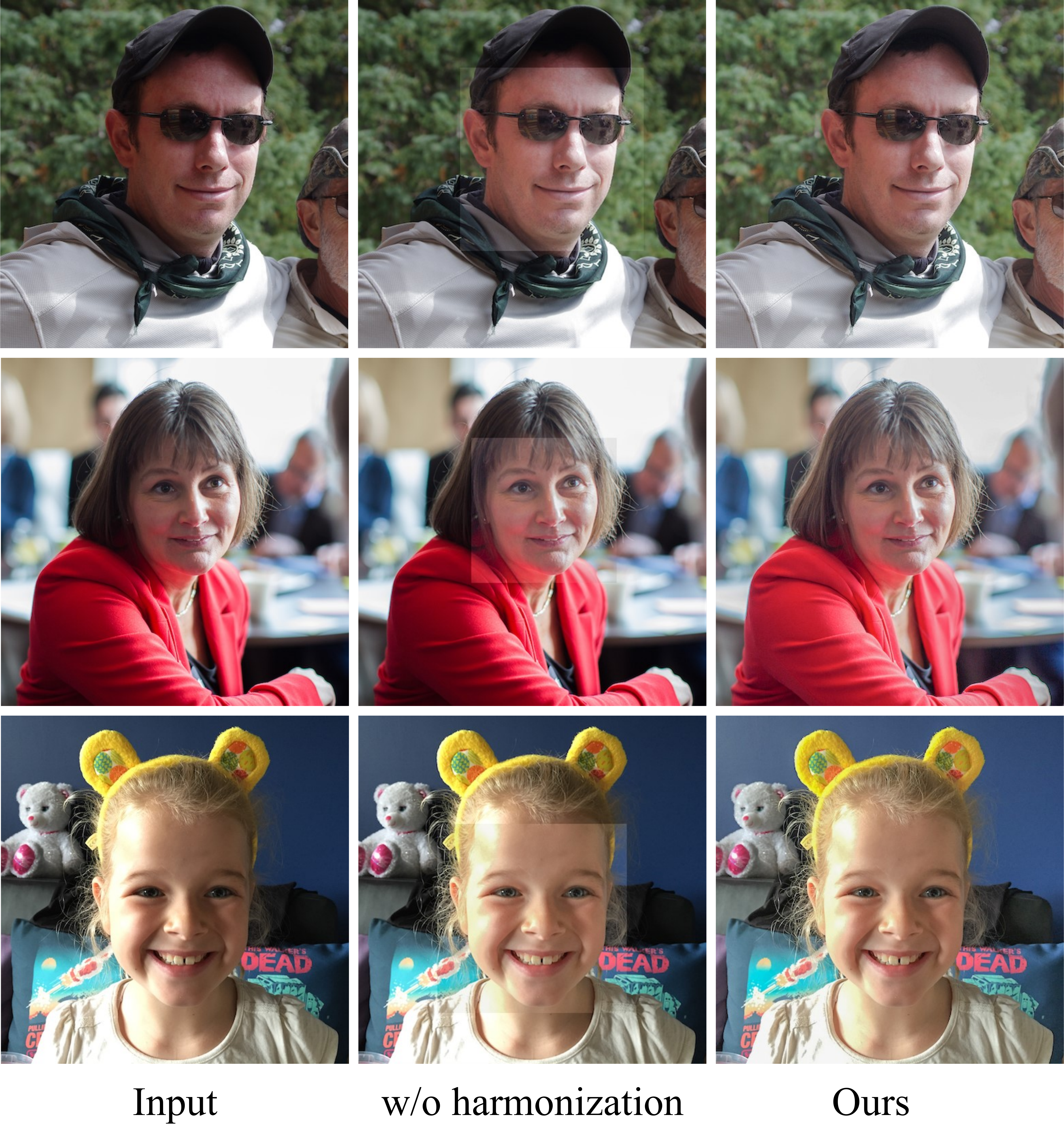}
   \caption{
     Image harmonization results. Three images of each sample are the input, the light-enhanced image without harmonization, and our result using harmonization respectively.}
    \label{fig:harm}
\end{figure}

To increase the stability for training the networks, we detect and crop the face region when pre-processing the dataset. To extend our method to handle images with random face size and location, 
we train an image harmonization network \cite{sofiiuk2020harmonization} to adjust the brightness of the background according to the enhanced face region.
In Figure \ref{fig:harm} we compare our harmonized enhancement results (Ours) with the inputs and the enhancement results directly padded to the original background (w/o harmonization). It can be seen that the discontinuity on the background or the hair region of the images after the enhancement is smoothed by the harmonization network, which effectively enriches the working scenarios of our method.

\begin{figure}[htb]
   \includegraphics[width=\linewidth]{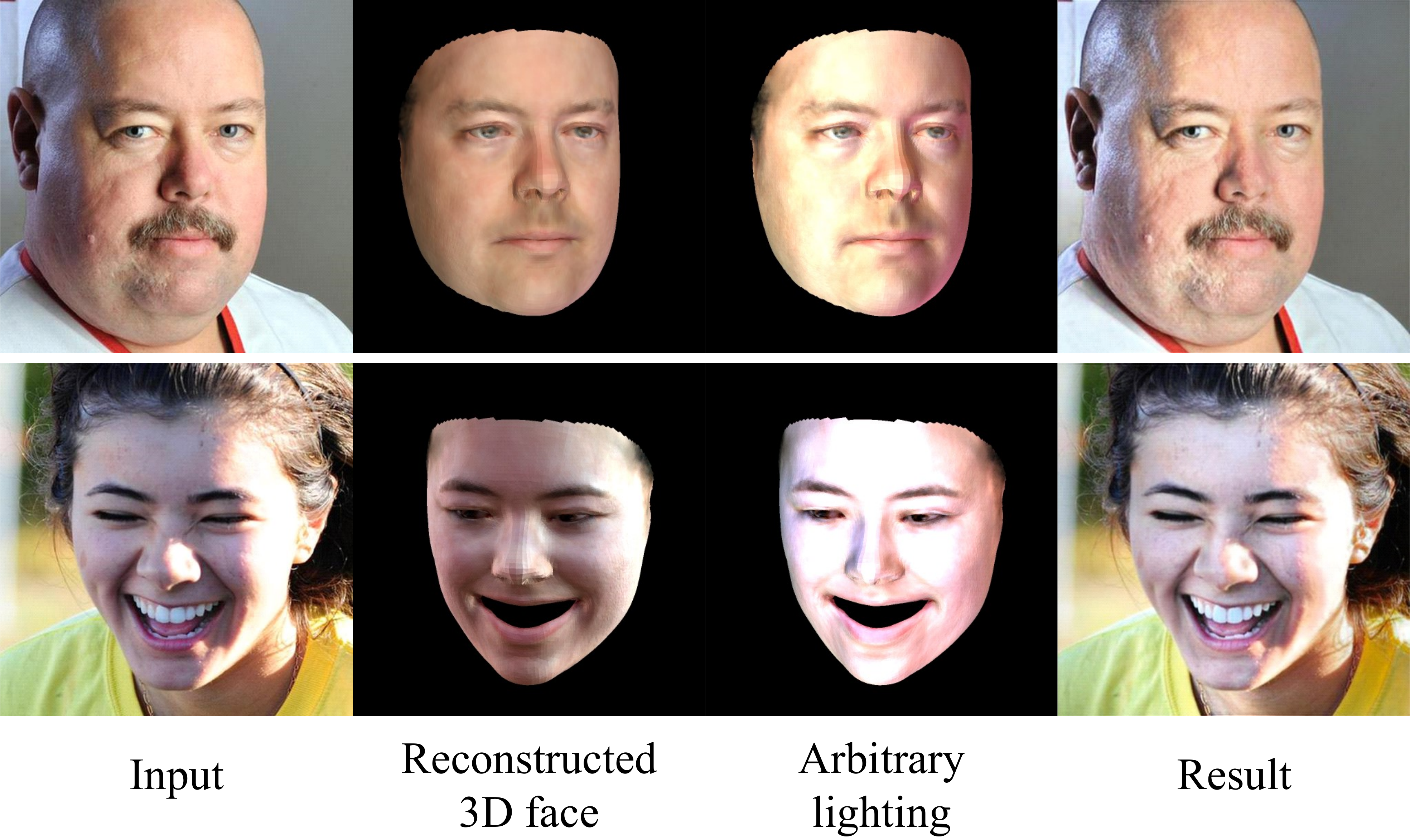}
   \caption{
     Failure cases when sampling the guidance with arbitrary lighting outside our enhanced lighting distribution.}
    \label{fig:limitation}
\end{figure}

\section{Conclusions}
In this paper, we present a deep face lighting enhancement method with 3D guidance. To support face light enhancement, we propose a method to synthesize more photo-realistic training pairs and use a 3DMM and a differentiable renderer to generate 3D guidance. We show that a transformer encoder can better model the relation between the guidance and the input low-light face image. 
Qualitative and quantitative experimental results prove the effectiveness of the proposed method in both poorly-lit synthetic and in-the-wild images.

Although it is proved that the lighting guidance is useful and more accurate and realistic guidance will lead to better enhancement performance, a limitation of current method is that it is unable to alter the lighting attributes, such as direction or strength, of the enhancement results with modifying the lighting guidance. This is because, unlike the relighting tasks where the lighting inputs (parameters or color maps) are only entangled with the training targets, the lighting guidance in our method also has correlation with the input image. Figure \ref{fig:limitation} shows some failure cases when providing the guidance with arbitrary lighting outside our enhanced lighting distribution. In future work, the framework can be modified to perform relighting function with generating multiple guidance-target pairs for one input image.
{Another potential limitation is, because a limited number of SH coefficients are used, the synthesized training pairs cannot represent all kinds of illuminations, especially those with high-frequency components. Thus, our network may not be able to handle these challenging illuminations perfectly. This is a common problem for all lighting enhancement and relighting methods with synthetic training data. While real datasets captured in light stages or real environments would be helpful to avoid such problems, they are expensive and time-consuming to collect.}

\section{Acknowledgements}
This work is supported by Huawei Ascend.

\bibliographystyle{eg-alpha-doi} 
\bibliography{refs}
\end{document}